\documentclass[final]{cvpr}

\usepackage{times}
\usepackage{epsfig}
\usepackage{graphicx}
\usepackage{amsmath}
\usepackage{bbm}
\usepackage{amssymb}

\usepackage{booktabs}
\usepackage{color}
\usepackage{amsfonts}
\usepackage{multirow}
\usepackage{mathtools}

\usepackage[pagebackref=true,breaklinks=true,colorlinks,bookmarks=false]{hyperref}

\def\cvprPaperID{2642} 
\def\confYear{CVPR 2021}

\begin{document}

\title{Learning Semantic-Aware Dynamics for Video Prediction}

\author{Xinzhu Bei\\
UCLA Vision Lab\\
{\tt\small xzbei@cs.ucla.edu}
\and
Yanchao Yang\\
Stanford University\\
{\tt\small yanchaoy@cs.stanford.edu}
\and
Stefano Soatto\\
UCLA Vision Lab\\
{\tt\small soatto@cs.ucla.edu}
}

\maketitle

\begin{abstract}
We propose an architecture and training scheme to predict video frames by explicitly modeling dis-occlusions and capturing the evolution of semantically consistent regions in the video.
The scene layout (semantic map) and motion (optical flow) are decomposed into layers, which are predicted and fused with their context to generate future layouts and motions. 
The appearance of the scene is warped from past frames using the predicted motion in co-visible regions; dis-occluded regions are synthesized with content-aware inpainting utilizing the predicted scene layout.
The result is a predictive model that explicitly represents objects and learns their class-specific motion, which we evaluate on video prediction benchmarks.
\end{abstract}

\section{Introduction}

\begin{figure}[!t]
    \centering
    \includegraphics[width=0.48\textwidth,trim=0 0 5 5, clip]{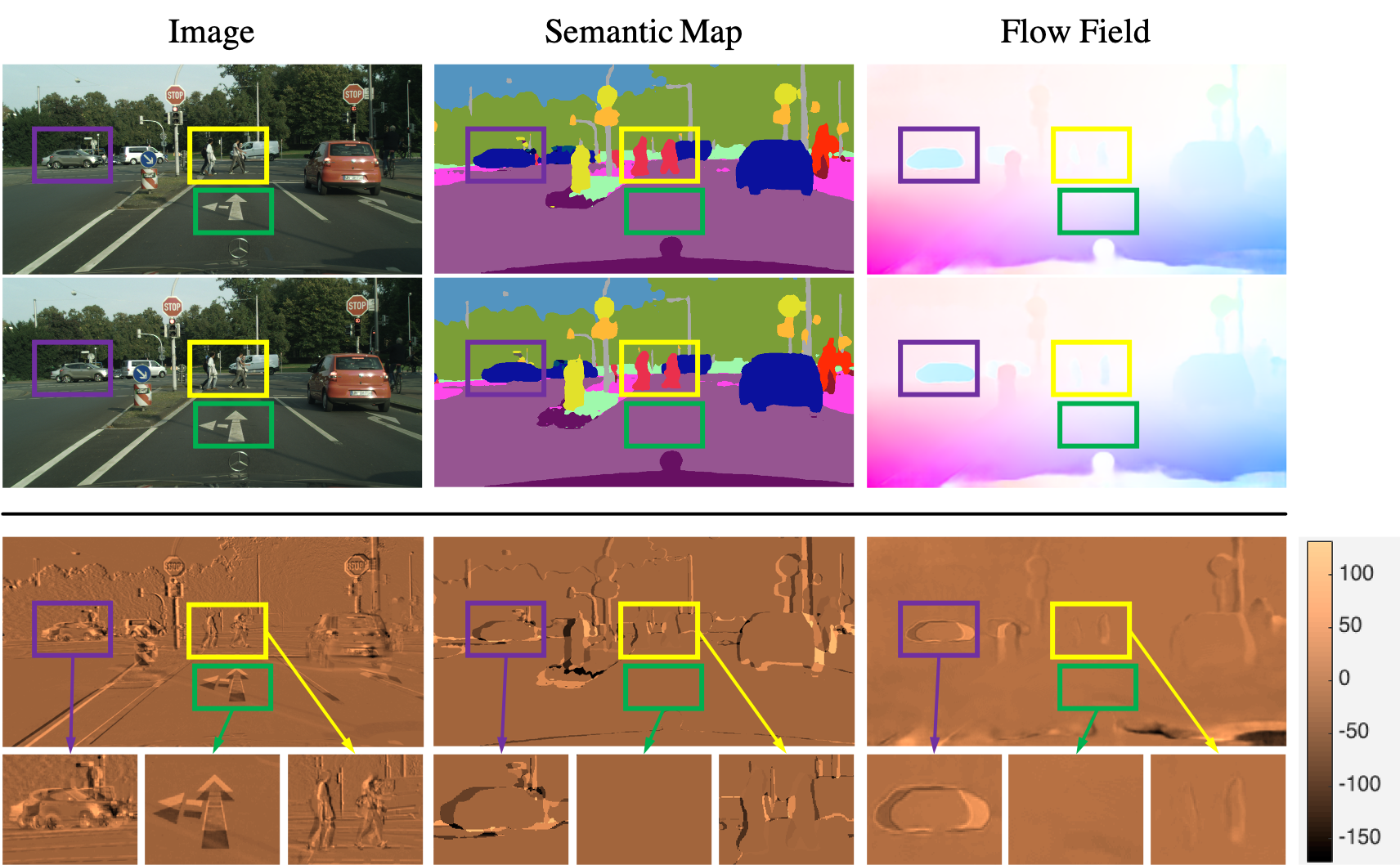}
    \caption{Different representations (video frame, semantic map, flow field) have dynamics with different complexity. Also, different classes have different dynamics within a given representation. Top: a sequence of video frames (left), semantic maps (middle), and flow fields (right). Bottom: dynamics or changes visualized in terms of their difference. The dynamics in video frames is much more complex than that in semantic maps and flow fields.}
    \label{fig:motivation}
\end{figure}

Anticipating the future is critical for autonomous agents to operate intelligently in the environment, such as for navigation, manipulation, and other forms of physical interaction. 
We hypothesize that decomposing the scene into independent entities, each with its own attributes, is beneficial to prediction.
For example, in Fig.~\ref{fig:motivation}, different objects have different geometry and motion, which induces distinctive temporal changes in the video.

\begin{figure*}[!t]
    \centering
    \includegraphics[width=0.9\textwidth]{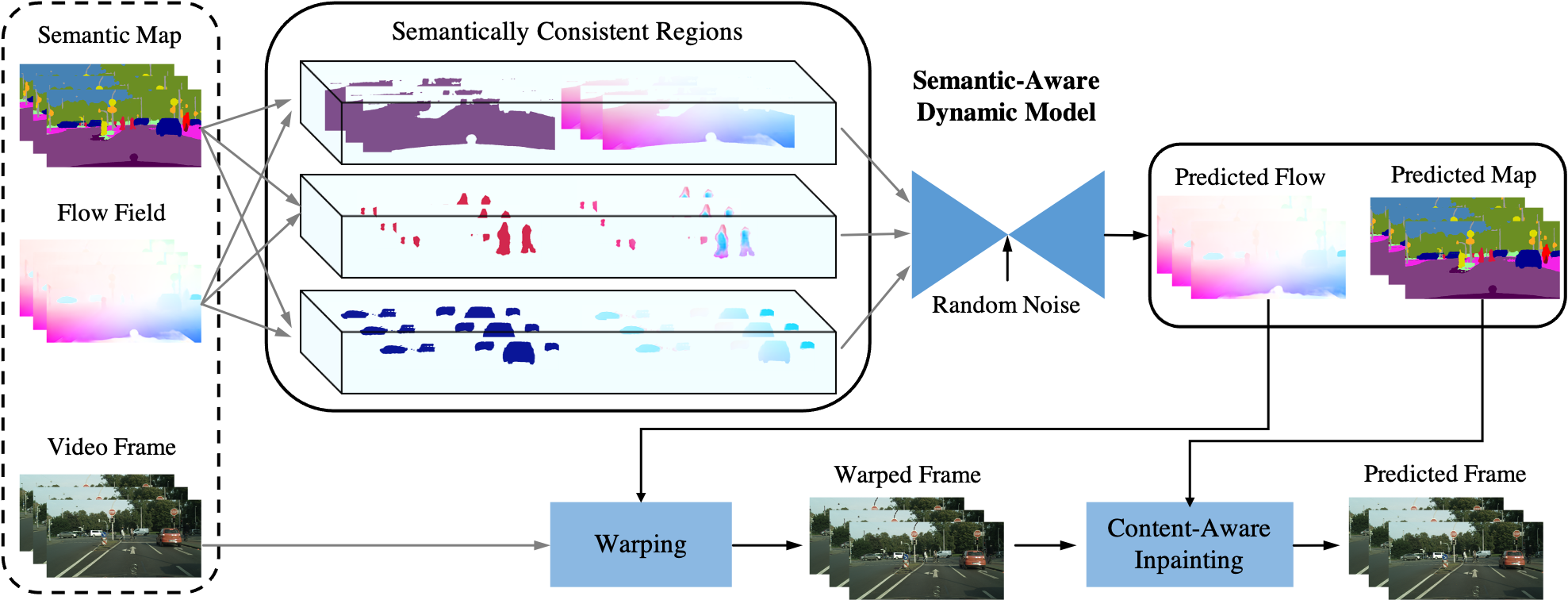}
    \caption{Our video prediction architecture with learned semantic-aware dynamics. It first decomposes the scene into semantically consistent regions to facilitate the modeling of class-specific characteristics. Each region is predicted and fused to generate the future scene layout (semantic map) and motion (flow field) using the proposed semantic-aware dynamic model. Content-aware video inpainting for dis-occlusions is performed after warping to generate the future video frames.}
    \label{fig:categorical_dynamics}
\end{figure*}

We propose a video prediction architecture that explicitly models the different dynamics of semantically consistent regions (Fig.~\ref{fig:categorical_dynamics}). The model, described in detail in Sec.~\ref{sec:sadm}, decomposes the video into regions, corresponding to different semantic classes in the scene, and learns class-specific characteristics while ensuring that their re-composition can predict the image, along with class labels and flow fields.

Unlike warping the past using globally predicted flow fields \cite{liang2017dual,li2018flow,pan2019video,gao2019disentangling}, in our semantic-aware dynamic model (SADM), local regions are represented by binary semantic masks, whose evolution is simpler and easier to learn than the motion of the entire video frames (see Fig. \ref{fig:motivation}). 
Each of the regions is predicted and then fused with its content to generate future semantic maps and flow fields.
The prediction in co-visible regions of future frames is warped from the past, with dis-occlusion detection mediated by the predicted semantic maps. 
Furthermore, the {\em dis-occluded} regions are filled-in by a generative model or conditional renderer, trained with not only the warped images, but also the predicted semantic maps, enabling more structured and semantically-aware synthesis. Modeling dis-occlusions explicitly spares the model the effort otherwise needed to learn  this complex phenomenon. 

We incorporate semantic segmentation (scene layout), optical flow (scene motion) and synthesis (scene appearance) into a complete generative model for videos, which facilitates semantically and geometrically consistent prediction of complete video frames. SADM achieves state-of-the-art performance in video prediction benchmarks such as \cite{Cordts2016Cityscapes,Geiger2012CVPR,Geiger2013IJRR}.

\section{Related Work}

\textbf{Video generation} methods produce image sequences either from noise \cite{vondrick2016generating} or other input including pose \cite{cai2018deep} and text \cite{marwah2017attentive}. 
SVG-LP \cite{denton2018stochastic} proposes to sample noise from learned priors; 
MoCoGAN \cite{tulyakov2018mocogan} samples latent variables from the motion and content spaces separately to improve temporal consistency. Similarly, TGAN \cite{saito2017temporal} employs a temporal generator and an image generator to model temporal correlations; \cite{he2018probabilistic} models the dynamics in the latent space with attribute controls. 
Given that the visual scene is highly structured, \cite{villegas2017learning,cai2018deep,yang2018pose} propose to generate a sequence of poses, which are transformed into images for human action sequences; \cite{zhao2018learning} generates videos of a single object by first generating a sequence of conditions using a 3D morphable model, while \cite{hao2018controllable} controls the video generation using sparse trajectories specified by the user. VGAN \cite{vondrick2016generating} trains video generators with explicit separation of the foreground and background, assuming static background. Seg2vid \cite{pan2019video} resorts to warping using flows generated by the semantic mask, hoping to preserve the scene structure implicitly. We also employ semantic maps in the generation of future flows but with a semantic-aware dynamic model. 
\cite{yang2020learning} decomposes images into objects utilizing contextual information separation \cite{yang2019unsupervised} and synthesizes motion of single objects through perturbations in the object-centric latent space.

\textbf{Video prediction} models are typically approximations of conditional generative models \cite{hsieh2018learning,babaeizadeh2017stochastic,wichers2018hierarchical,vondrick2016generating,saito2017temporal,denton2018stochastic,tulyakov2018mocogan,reda2018sdc,xu2018structure}. The quality of predictions is typically evaluated by image quality and temporal consistency. Given the high complexity and dimensionality of the signal to be predicted, the process usually requires explicit modeling or constraints \cite{pan2019video,gao2019disentangling}. PredNet \cite{lotter2016deep} proposes a predictive model with coding-based regularization. ContextVP \cite{byeon2018contextvp} uses a context-aware module with parallel LSTMs. SDC-Net \cite{reda2018sdc} applies flow guided spatially-displaced convolutions, while \cite{jia2016dynamic} predicts with dynamic filters that depend on the inputs. DDPAE \cite{hsieh2018learning} and \cite{wichers2018hierarchical} map the observed images to a low-dimensional space, so temporal correlations are easier to learn. TPK \cite{walker2017pose} predicts future poses to guide appearance changes. To address the loss of realism, \cite{pintea2014deja,liang2017dual,luo2017unsupervised,li2018flow} explicitly model the flows, and DVF \cite{liu2017video} uses flow to synthesize future frames.
Similar to MCNet \cite{villegas2017decomposing} and \cite{xue2016visual}, DPG \cite{gao2019disentangling} proposes motion-specific propagation and motion-agnostic generation with confidence-based occlusion maps. 
\cite{luc2017predicting} predicts future semantic maps, and \cite{jin2017predicting} jointly predicts the future semantic maps and flow fields. We use a semantic-aware model such that the predicted maps can exploit class-specific motion priors. Our method generates both future optical flows and semantic maps before rendering future images.
In \cite{wu2020future}, moving object segmentation masks are used, but restricted to 2D affine motions, with two categories: moving and static.

\textbf{Image inpainting} \cite{patwardhan2007video,xu2019deep,kim2019deep}, image synthesis \cite{brock2018large,reed2016generative,wang2018high}, and video-to-video synthesis \cite{wang2018video} are also related to our handling of dis-occlusions.

\section{Method}

\begin{figure*}[!t]
    \centering
    \includegraphics[width=0.95\textwidth]{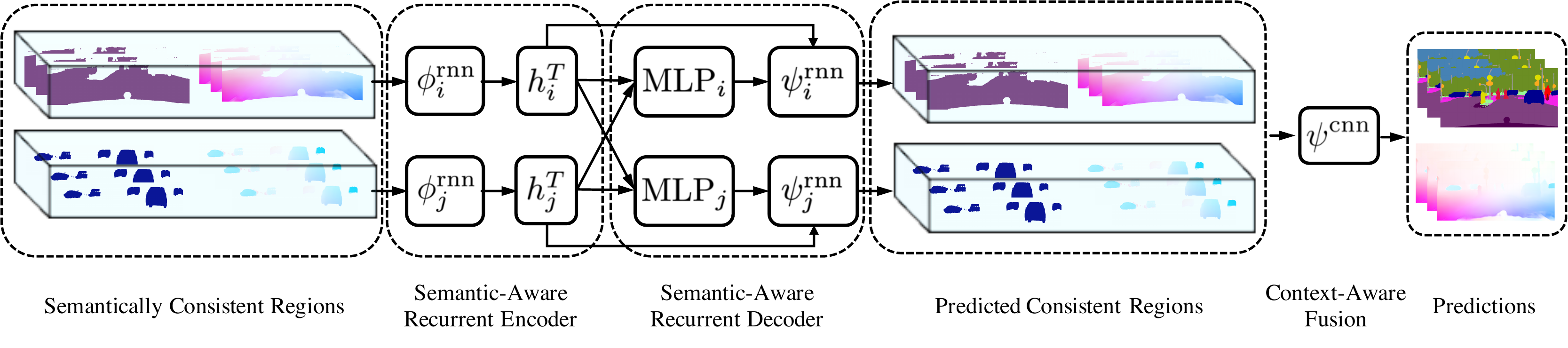}
    \caption{The architecture of our semantic-aware dynamic model (SADM) for learning class-specific dynamics of the scene layout and motion. 
    Input semantic maps and flow fields are parsed and processed by the semantic-aware recurrent encoders $\phi^{\mathrm{rnn}}$ and decoders $\psi^{\mathrm{rnn}}$, with context incorporated into the prediction through a multi-layer perceptron.
    The predictions of semantically consistent regions are combined by the fusion network $\psi^{\mathrm{cnn}}$ to generate the prediction on the whole image domain, which further improves contextual compatibility in the predicted semantic maps and flow fields.
    The illustration is for two classes, but can be easily extended to more classes.}
    \label{fig:architecture}
\end{figure*}

\noindent {\bf Notation and goal.} 
Let $x^t \in \mathbb{R}^{H\times W\times 3}$ be a video frame at time
$t$, with $f^t \in \mathbb{R}^{H\times W\times 2}$ and $m^t \in \{1,2,...,C\}^{H\times W}$ be the corresponding optical flow field and semantic map respectively.
Here $C$ is the number of semantic classes in the semantic map.
Given past observations $\{x^t, f^t, m^t\}_{t=1}^T$ up to time $T$, 
our goal is to predict $K$ frames into the future, i.e., $\{x^t\}_{t=T+1}^{T+K}$. 
The predictions should match the statistics, quality and content of past frames of the same scene, and exhibit variations that are consistent with the motion of objects within.
As illustrated in Fig.~\ref{fig:categorical_dynamics}, our approach falls into the direction of prediction by propagation, where video prediction for the co-visible part\footnote{image regions that are observed/visible across multiple frames} of the scene can be accomplished by warping, i.e., propagating pixels via the corresponding flow field.

\subsection{Semantic-Aware Dynamic Model}
\label{sec:sadm}

We aim to explicitly model the semantic aware dynamics of both the semantic maps (scene layout) and flow fields (scene motion).
In general, the proposed semantic aware dynamic model takes as input the flow fields and semantic maps up to time $T$, i.e., $\{(m^t, f^t)\}_{t=1}^{T}$ and outputs the $K$ future flow fields and semantic maps $\{(m^t, f^t)\}_{t=T+1}^{T+K}$, as shown in Fig.~\ref{fig:architecture}. 
The components therein are elaborated below.

\noindent {\bf Semantic-aware recurrent encoder.} Let $m^t_c = \mathbbm{1}(m^t = c)$ 

be the binary mask that indicates the region of semantic class $c$ at time $t$. Similarly, $f^t_c = f^t \cdot m^t_c$ is the masked flow field showing only the motion of the pixels that are classified as $c$.
The semantic aware recurrent encoder $\phi_c^{\mathrm{rnn}}$ will be operating recursively to produce a hidden representation of the past $\{(m_c^t,f_c^t)\}_{t=1}^T$ while enforcing temporal continuity of the representation:
\begin{equation}
    h^t_c = \phi_c^{\mathrm{rnn}}( [m_c^t,f_c^t], h^{t-1}_c)
\end{equation}
with $h^T_c$ the hidden representation that summarizes the past regions and flow fields of the pixels within class $c$, up to time $T$.
For now, we instantiate $C$ such semantic aware recurrent encoders, $\{\phi_c^{\mathrm{rnn}}\}_{c=1}^C$, which together generate the hidden representation $H^T = \{h^T_c\}_{c=1}^{C}$ that summarizes the past semantic maps and flow fields, covering all semantic classes. 

Note that the collection $H^T$ explicitly represents independent objects. 
While this may appear inefficient, in reality the model reduces the number of parameters needed, since the individual objects are simpler to represent. 
We also carry out an ablation study (in the supplementary Sec. \ref{sec:supp_ablation}) on different $C$'s by merging some of the semantically similar classes, showing the accuracy-efficiency trade-offs. 
Moreover, we can easily parallelize the computation using the grouped convolution operator proposed in \cite{krizhevsky2017imagenet}. Next, we describe the procedure to predict the future semantic maps and flow fields.

\noindent {\bf Semantic aware recurrent decoder.} 
Given the hidden representation of the past, $H^T = \{h^T_c\}_{c=1}^{C}$, 
the semantic aware recurrent decoder produces $K$ future semantic maps and flow fields $\{(m^t, f^t)\}_{t=T+1}^{T+K}$. 
We first describe a deterministic decoding procedure, for simplicity, which can then be easily adapted to a stochastic one to account for the randomness of the future.

Again, we consider decoders that learn the dynamics and predict the future in a semantic aware manner. Let $\psi^{\mathrm{rnn}}_c$ be the recurrent decoder for semantic class $c$, which generates the prediction for $\{(m^t_c,f^t_c)\}_{t=T+1}^{T+K}$ by recursively executing the following procedures:
\begin{align}
    h^t_c, e^t_c &= \psi^{\mathrm{rnn}}_c( h^{t-1}_c, \mathrm{MLP}_c(H^T) ), t \geq T+1 \label{eq:decode-rnn} \\ 
    \Tilde{m}^t_c &= \psi^{\mathrm{rnn}}_{c,m}(e^t_c); \quad
    \Tilde{f}^t_c = \psi^{\mathrm{rnn}}_{c,f}(e^t_c)
\end{align}
Here we abuse the notation $\psi^{\mathrm{rnn}}_c$ to refer to the recurrent unit that updates the latent representation $h^t_c$, while generating a common embedding $e^t_c$, which is then decoded into the predicted semantic mask $\Tilde{m}^t_c$ and flow fields $\Tilde{f}^t_c$, respectively through separate decoding heads $\psi^{\mathrm{rnn}}_{c,m}$ and $\psi^{\mathrm{rnn}}_{c,f}$. 
This separate decoding design aligns with the practice that improves the decoding efficiency in multi-task learning. 
Note, we also apply a multi-layer perceptron $\mathrm{MLP}_c$ (due to its efficiency) on the collection of the hidden representations for all classes $H^T = \{h^T_c\}_{c=1}^{C}$, to ensure that the semantic aware decoder has access to the context provided by other classes within the scene (Fig.~\ref{fig:architecture}).

The decoders for each class $\{\psi^{\mathrm{rnn}}_{c,m}, \psi^{\mathrm{rnn}}_{c,f}\}_{c=1}^C$ can also be running in parallel, so that we have the semantic aware predictions for each class, 
i.e., $\{(\Tilde{m}^t_c,\Tilde{f}^t_c)\}$ with $t\in \{T+1,...,T+K\}$ and $c\in \{1,...,C\}$.
Next, we apply late fusion to get predictions that can be directly compared to the ground-truth semantic maps and flow fields $\{(m^t,f^t)\}_{t=T+1}^{T+K}$, and to further improve the contextual compatibility between different classes.

\noindent {\bf Context-aware late fusion.}
Given $\{(\Tilde{m}^t_c,\Tilde{f}^t_c)\}_{t=T+1}^{T+K}$ for each $c\in \{1,...,C\}$, we apply a three-layer ConvNet to first fuse the binary semantic maps:
\begin{equation}
    \Tilde{m}^t = \psi^{\mathrm{cnn}}(\mathrm{concat}(\{\Tilde{m}^t_c\}_{c=1}^C))
\end{equation}
where the dimension of $\Tilde{m}^t$ is $H \times W \times C$. We use softmax as the last layer for $\psi^{\mathrm{cnn}}$, such that each slice of $\Tilde{m}^t$ indexed by the last dimension, i.e., $\Tilde{m}^t(c) = \Tilde{m}^t[:,:,c]$ (in Python style), is still a scalar field indicating the probability of each pixel belonging to class $c$.
And the fused flow field $f^t$ is obtained as following:
\begin{equation}
    \Tilde{f}^t(i,j) = \sum_c \Tilde{m}^t(i,j,c)\cdot \Tilde{f}^t_c(i,j);\ \  i \leq H, j \leq W
\end{equation}
which is a linear combination of the flow vectors predicted by each semantic aware recurrent decoder, whose visibility comes from the fused semantic map $\Tilde{m}^t$.

\noindent {\bf Training loss for $\phi^{\mathrm{rnn}}_c, \mathrm{MLP}_c, \psi^{\mathrm{rnn}}_c, \psi^{\mathrm{rnn}}_{c,m}, \psi^{\mathrm{rnn}}_{c,f}, \psi^{\mathrm{cnn}}$.} With the ground-truth $\{(m^t,f^t)\}_{t=T+1}^{T+K}$, the training loss for flow fields is the $L1$ loss:
\begin{equation}
    \mathcal{L}_f = \sum_{t=T+1}^{T+K} \| \Tilde{f}^t - f^t\|_1
\end{equation}
which penalizes the discrepancy between the predicted flow and the ones computed from the ground-truth images. For the semantic maps we apply the cross entropy loss:
\begin{equation}
    \mathcal{L}_m = \sum_{t=T+1}^{T+K} (1+\alpha\ \mathcal{G}*\nabla m^t) \cdot \mathcal{H}(m^t, \Tilde{m}^t)
\end{equation}
Here $\mathcal{H}$ represents the cross-entropy, which is weighted by whether the pixel is near the boundaries between different classes or not. 
Note $\nabla$ is the gradient operator, and we binarize its response to 0 and 1 to discount the artifacts caused by naming different classes with different integers. 
The binarized boundary map is then smoothed by a Gaussian kernel $\mathcal{G}$ to expand the weights to nearby pixels, making the boundaries thicker. The variance of the Gaussian, which determines the spatial extent of the boundaries is set to $9.0$ and fixed.
With this weighting scheme, the network will focus more on the pixels near the semantic boundaries, thus better preserves the shape of each semantic segment in the prediction. The relative importance between boundary and non-boundary pixels is controlled by the scalar $\alpha$, which is set to $5.0$ for all experiments.

So far, we have described the proposed semantic aware dynamic model in its deterministic mode. 
However, extending it to account for the stochasticity of the future is straight-forward.
For this purpose, we instantiate $C$ semantic aware recurrent encoders $\theta^{\mathrm{rnn}}_c$, which operate in a similar way as the encoders for the past:
\begin{equation}
    z^t_c = \theta^{\mathrm{rnn}}_c([m^t_c,f^t_c], z^{t-1}_c);\ \ z^t_c = [u^t_c,v^t_c],\ t \geq T+1
\end{equation}
The goal of the recurrent encoder $\theta^{\mathrm{rnn}}_c$ is to generate a random variable $z^t_c$, represented by its mean and variance $[u^t_c, v^t_c]$ through reparameterization, whose initial value is set to $z^T_c = [h^T_c, I]$.\footnote{this ensures that the generation of the random variable is conditioned on the past.} 
At the end of the recursion, we would like $z^{T+K}_c = [u^{T+K}_c, v^{T+K}_c]$ to be a zero-mean unit-variance Gaussian.
Then, $Z^K = \{(u^{T+K}_c,v^{T+K}_c)\}_{c=1}^C$ will be added to $H^T$ in Eq.~\eqref{eq:decode-rnn}, also through reparameterization, for decoding the future with randomness.
To learn $\theta^{\mathrm{rnn}}_c$'s, we add a KL-divergence term to the loss:
\begin{equation}
    \mathcal{L}_{\text{kl}} = \mathbb{KL}(\mathcal{N}(u^{T+K}, v^{T+K}), \mathcal{N}(\mathbf{0}, \mathbf{I}))
\end{equation}
where $\mathcal{N}$ represents the normal distribution. 
We summarize the training loss for the stochastic semantic aware dynamic model in the following:
\begin{equation}
    \mathcal{L}_{\text {dynamic}} = \mathcal{L}_f + \mathcal{L}_m + \beta \mathcal{L}_{kl}
    \label{eq:vae-loss}
\end{equation}
with $\beta$ the weight on the KL-divergence term. As in VAEs, $\{\theta^{\mathrm{rnn}}_c\}_{c=1}^C$ are used only during the training for the stochastic decoder, and will not be used during testing since the random noise can be directly sampled from the prior $\mathcal{N}(\mathbf{0}, \mathbf{I})$.

\begin{figure}[!t]
    \centering
    \includegraphics[width=0.47\textwidth]{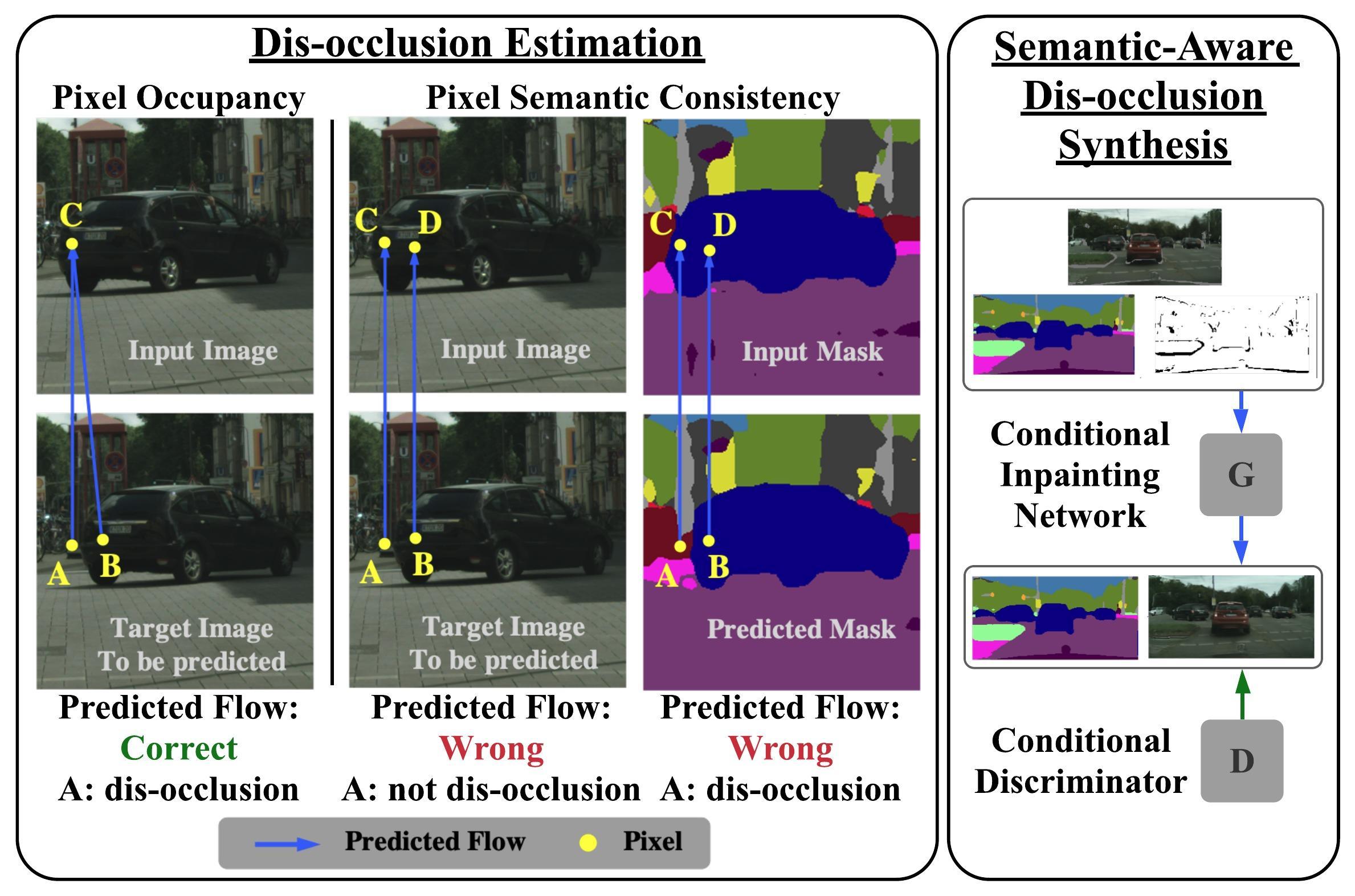}
    \caption{\textbf{Left}: Two criteria for dis-occlusion detection. {\em Pixel Occupancy}: pixels A, B in the target image domain are mapped onto pixel C in the input image domain, which is occupied by more than one pixel when the predicted (backward) flow (blue arrows) is correct; in this case, pixel A, as the cause of over-occupancy, can be detected as dis-occlusion. {\em Pixel Semantic Consistency}: if the predicted flow is incorrect, pixel occupancy fails in detecting A as dis-occlusion; however, given that semantic mask can be accurately predicted, A will be mapped to pixel C with inconsistent semantic labels, thus can be correctly detected as dis-occlusion. \textbf{Right}: Semantic-aware dis-occlusion synthesis, where both the generator and the discriminator take in predicted semantic mask, and the generator is dis-occlusion aware.}
    \label{fig:occu_warp}
\end{figure}

\subsection{Warping with Semantic Informed Dis-occlusion}

We warp the past video frames to provide an anchor point for future synthesis using the predicted future semantic masks and flows fields $\{(\Tilde{m}^t,\Tilde{f}^t)\}_{t=T+1}^{T+K}$ from the semantic-aware dynamic model.
To ease the warping and comply with the literature, here we mark the predicted flow $\Tilde{f}^t$ as the backward flow, i.e., from $t+1$ to $t$. 
The warping can be simply performed via bilinear interpolation:
\begin{equation}
    \hat{x}^{t+1}(p) = x^t(p+\Tilde{f}^{t+1}(p)), p \notin \Omega_d
\end{equation}
The key is to estimate the dis-occluded area $\Omega_d$, which invalidates the assumption that a pixel in frame $x^{t+1}$ is propagated from the previous frame $x^t$.

Note, \cite{gao2019disentangling} proposes to use pixel occupancy for dis-occlusion detection, however, miss-detection happens due to errors in flow prediction on the object boundaries where dis-occlusion resides (see Fig.~\ref{fig:occu_warp}).
Given that semantic masks are easier to predict than flows, particularly, with our semantic-aware dynamic model, we propose a semantic consistency criteria for dis-occlusion estimation, i.e., $p \in \Omega_d, \text{if } \Tilde{m}^{t+1}(p) \neq m^t(p)$.
The above semantic consistency criteria can still correctly detect dis-occlusions even if the flow is wrong as shown in Fig.~\ref{fig:occu_warp}.
In our experiments, we use both the pixel occupancy and the proposed semantic consistency criterion for dis-occlusion detection given their complementarity. 

After warping, we end up with the future frames warped from the past and the corresponding dis-occlusion masks, i.e., $\{(\hat{x}^t, \Omega_d^t)\}_{t=T+1}^{T+K}$.
Note $\hat{x}^t$ is only valid (up to noise) in the complement of $\Omega_d^t$, which will be extrapolated as we describe next.

\subsection{Semantic-Aware Dis-occlusion Synthesis}

\begin{figure*}[t]
    \centering
    \includegraphics[width=0.93\textwidth]{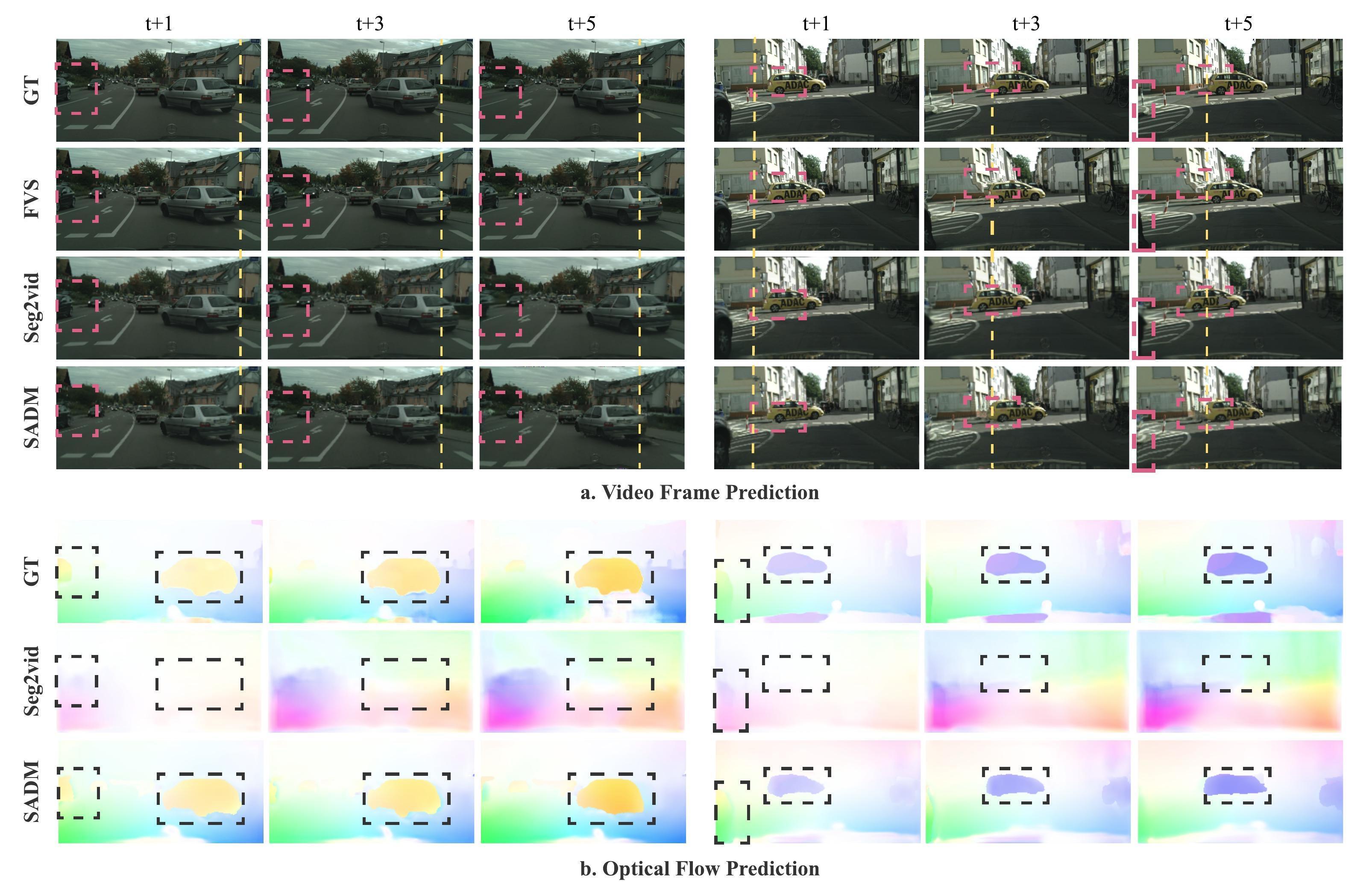}
    \caption{Visual comparison on the Cityscapes dataset. Both the predicted video frames (a) and flow fields (b) are presented. Left: the flow predicted by our network clearly shows the silver car moves to the left and the camera moves forward, while the flow predicted from Seg2vid \cite{pan2019video} is dominated by ``major" camera motion exhibited in the dataset, i.e., zooming-in caused by the movement of the running car. FVS \cite{wu2020future} wrongly predicts the motion of the silver car, resulting in incorrect car locations at $t+3$ and $t+5$. Note that, at $t+5$, the black car on the left should move outside the image domain, which is only captured by our model. Again, the ``ghost effect'' presents near the objects' boundaries in the predictions from the other two methods. Right: without conditioning on semantic segmentation masks, the dis-occluded area of the building is incorrectly inpainted by the inpainting network as part of the moving car, causing distortions in the results from FVS.}
    \vspace{-0.2cm}
    \label{fig:cityscape}
\end{figure*}

Using the warped frames $\{\hat{x}^{t}\}_{t=T+1}^{T+K}$ as the anchor, we employ a conditional inpainting network to further complete the dis-occluded parts and improve the quality of the synthesized images via adversarial training.
The conditional inpainting network $\varphi$ takes as input the anchor frame $\hat{x}^t$, and tries to complete the missing region indicated by $\Omega_d^t$ based on the predicted semantic map $\Tilde{m}^t$ in a content-aware manner:
\begin{equation}
    \Tilde{x}^t = \varphi(\hat{x}^t, \Omega_d^t, \Tilde{m}^t)
\label{eq:inpainting}
\end{equation}
Note, image details and their temporal consistency can be improved by semantic maps informing the scene content as shown in \cite{wang2018video}, which only focuses on translating known semantic maps to images.
Given the ability to model the dynamics of the semantic map and its prediction, 
our conditional inpainting network can be informed about the scene content, thus able to generate better synthesis (see Fig. \ref{fig:cityscape}).

To help the training of the content-aware conditional inpainting, we also employ two discriminators $D_v$, $D_x$ for the video clip and frame respectively, with $D_v$ focuses on the temporal continuity and $D_x$ focuses on the image quality.
So the training loss for the content-aware inpainting network is:
\begin{multline}
    \mathcal{L}_{\varphi} = \sum_{t=T+1}^{T+K} (1-\Omega_d^t)\cdot\|\Tilde{x}^t - \hat{x}^t\|_{1} + 
    \lambda \mathcal{L}_{per}( \Tilde{x}^t, x^t) \\
    + \gamma \sum_{t=T+1}^{T+K}  D_x(\Tilde{x}^t,\Tilde{m}^t) + \eta D_v(\{ \Tilde{x}^t \}, \{\Tilde{m}^t\}) 
\label{eq:inpainting-loss}
\end{multline}
where the first term measures the discrepancy between the completed image and the warped image in the co-visible area; the loss $\mathcal{L}_{per}$ measures the perceptual similarity between the generated images and the real images \cite{pan2019video}. $D_x$ and $D_v$ measure the plausibility of images/videos conditioned on the semantic content.
The final predictions of our model are $\{(\Tilde{m}^t, \Tilde{f}^t, \Tilde{x}^t)\}_{t=T+1}^{T+K}$, i.e., predicted semantic maps, flow fields and video frames with the semantic-aware dynamics model as their driving force.

\section{Experiments}
\label{sec:experiments}

{\bf Datasets.} We evaluate our method on multiple {\em prediction} tasks, e.g., video frames and semantic maps, using three commonly used datasets, Cityscapes \cite{Cordts2016Cityscapes}, KITTI Flow \cite{Geiger2012CVPR} and KITTI Raw \cite{Geiger2013IJRR}.
Cityscapes \cite{Cordts2016Cityscapes} contains driving sequences recorded in 50 different cities. 
We use the training split for training our semantic-aware dynamic model and the validation set for evaluation. 
The training and evaluation subsets contain 2975 and 500 videos, respectively. 
Pixel-wise annotations for semantic segmentation are only available every 20 frames for the Cityscapes dataset. 
KITTI Raw \cite{Geiger2013IJRR} contains 156 long sequences. 
Following \cite{wu2020future}, we use 4 of them for testing, and the rest for training. 
KITTI Flow \cite{Geiger2012CVPR} is designed for benchmarking optical flow algorithms, and is more challenging than KITTI Raw \cite{Geiger2013IJRR}. 
It consists of 200 training videos and 200 test videos. 
Following \cite{gao2019disentangling}, we downsample the videos to $128 \times 424$ and then center-crop to $128 \times 256$, 
yielding 4000 clips for both training and testing.
Since per-frame dense semantic maps and optical flow annotations are not available, we leverage the off-the-shelf semantic segmentation network DeepLabV3 \cite{chen2018encoder} to extrapolate annotations for 20 classes, and compute the optical flow using the PWC-Net \cite{sun2018pwc}.

{\bf Implementation and training.} We adapt the grouped Conv-LSTM network \cite{xingjian2015convolutional} for the semantic-aware recurrent encoders and decoders to perform temporal aggregation of semantic maps and flow fields. 
The inpainting network is a modified U-Net \cite{ronneberger2015u}, conditioned on predicted anchor frames and semantic maps. 
We also replace the convolutional layers in the encoder with the partial convolution proposed in \cite{liu2018image}, which masks dis-occluded area in the feature space at different resolutions. 
The video and image discriminators are similar to those in CycleGAN \cite{CycleGAN2017}, except that the video discriminator has ordinary 2D convolutions replaced with 3D convolutions.

Though our model can be trained end-to-end, we split the training into two stages to manage on a single work-station:
1) Training the semantic-aware dynamic model with loss \eqref{eq:vae-loss}; 2) Training the inpainting network to fill-in the dis-occluded area with loss  \eqref{eq:inpainting-loss}. 
The training is performed on 2 GeForce GTX 1080 Ti GPUs with batch size equals to 6, and each video clip in the batch contains 10 frames. 
Learning rates for both stages start from $0.001$ and decay by 0.8 every 20 epochs.
The weights of each term in the losses are detailed in the supplementary (Sec. \ref{sec:supp_weights}).
The training of the semantic aware dynamic model needs 40 hours to converge, and the training of the inpainting network takes about 20 hours. 

\textbf{Model complexity and inference.} The semantic-aware dynamic model for predicting semantic maps and flow fields contains 8.6M parameters. The inpainting network contains 5.18M parameters. Inference can be performed on a single GeForce GTX 1080 GPU with 8GB memory. The inference runs at 19 frames per second.

\begin{figure*}[t]
    \centering
    \includegraphics[width=0.93\textwidth]{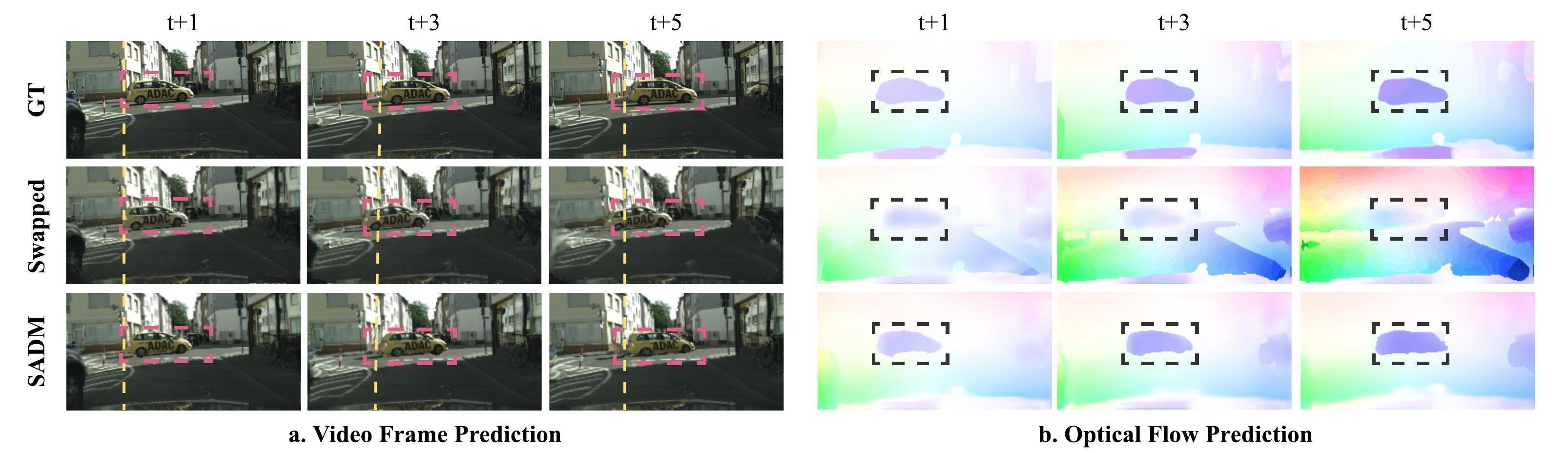}
    \caption{Ablation study: our model predicts video frames and flow fields by swapping semantic-aware encoders for the classes of ``car'' and ``road'', verifying that semantic-aware dynamics is learned with the proposed model.}
    \vspace{-0.3cm}
    \label{fig:permute}
\end{figure*}

\subsection{Quantitative Results}

\textbf{Video prediction.} Table \ref{table:cityscape} and \ref{table:kitti_raw} report the multi-frame video prediction performance evaluated in terms of Multi-scale Structural Similarity Index Measure (MS-SSIM) \cite{wang2003multiscale} and LPIPS \cite{zhang2018unreasonable}, on the Cityscapes and the KITTI Raw datasets respectively. 
Higher MS-SSIM scores and lower LPIPS distances suggest better performance. 
Specifically, on longer horizon prediction ($t+5$), our model improves Seg2vid \cite{pan2019video}, which also employs semantic segmentation, by 35.50\% (MS-SSIM) and 20.85\% (LPIPS). 
Moreover, our model outperforms FVS \cite{wu2020future}, which infers 2D affine transformations of moving objects, by 10.35\% (MS-SSIM) and 9.39\% (LPIPS) on the $t+5$ predictions.

\begin{table}[t]
\setlength{\tabcolsep}{4pt}
\centering
\small
\begin{tabular}{ccccc}
\toprule
    & \multicolumn{2}{c}{\textbf{MS-SSIM} ($\times$1e-2) $\uparrow$} & \multicolumn{2}{c}{\textbf{LPIPS} ($\times$1e-2) $\downarrow$} \\
    \textbf{Method} & t+1 & t+5 & t+1 & t+5                \\
    \midrule
    PredNet \cite{lotter2016deep} & 84.03 & 75.21             & 25.99             & 36.03             \\
    MCNET \cite{villegas2017decomposing}      & 89.69             & 70.58             & 18.88             & 37.34             \\
    Voxel Flow \cite{liu2017video} & 83.85             & 71.11             & 17.37             & 28.79             \\
    Vid2vid \cite{wang2018video}    & 88.16             & 75.13             & 10.58             & 20.14             \\
    Seg2vid \cite{pan2019video}    & 88.32             & 61.63             & 9.69             & 25.99             \\
    FVS \cite{wu2020future}        & 89.10             & 75.68             & 8.50              & 16.50              \\
    \midrule
    SADM & \textbf{95.99}    & \textbf{83.51}    & \textbf{7.67}    & \textbf{14.93}    \\
    \bottomrule
\end{tabular}
\vspace{0.1cm}
\caption{Quantitative comparison on the Cityscapes dataset.}
\vspace{-0.2cm}
\label{table:cityscape}
\end{table}

\begin{table}[h]
\centering
\small
\setlength\tabcolsep{4.0pt}
\begin{tabular}{ccccccc}
\toprule
& \multicolumn{3}{c}{\textbf{MS-SSIM} ($\times$1e-2) $\uparrow$} & \multicolumn{3}{c}{\textbf{LPIPS} ($\times$1e-2) $\downarrow$} \\
\textbf{Method}     & t+1         & t+3         & t+5         & t+1      & t+3      & t+5               \\
\midrule
PredNet \cite{lotter2016deep}    & 56.26      & 51.47      & 47.56      & 55.35   & 58.66   & 62.95            \\
MCNet \cite{villegas2017decomposing}      & 75.35      & 63.52      & 55.48      & 24.05   & 31.71   & 37.39            \\
Voxel Flow \cite{liu2017video} & 53.93      & 46.99      & 42.62      & 32.47   & 37.43   & 41.59            \\
FVS \cite{wu2020future}        & 79.28      & 67.65      & 60.77      & 18.48   & 24.61   & \textbf{30.49}   \\
\midrule
SADM & \textbf{83.06}      & \textbf{72.44}      & \textbf{64.72}      & \textbf{14.41}   & \textbf{24.58}   & 31.16            \\
\bottomrule
\end{tabular}
\vspace{0.0cm}
\caption{Quantitative comparison on the KITTI Raw dataset.}
\vspace{-0.2cm}
\label{table:kitti_raw}
\end{table}

\begin{table}[h]
\centering
\small
\setlength\tabcolsep{3.0pt}
\begin{tabular}{cccc}
\toprule
 \textbf{Method} & \textbf{PSNR}$\uparrow$           & \textbf{SSIM} ($\times$1e-2)$\uparrow$ & \textbf{LPIPS} ($\times$1e-2)$\downarrow$ \\
 \midrule
 Repeat \cite{gao2019disentangling} & 16.5           & 48.9              & 19.0              \\
 PredNet \cite{lotter2016deep}       & 17.0           & 52.7              & 26.3              \\
 SVP-LP \cite{denton2018stochastic}   & 18.5           & 56.4              & 20.2              \\
 MCNet \cite{villegas2017decomposing} & 18.9           & 58.7              & 23.7              \\
 MoCoGAN \cite{tulyakov2018mocogan}   & 19.2           & 57.2              & 18.6              \\
 DVF \cite{liu2017video}              & 22.1           & 68.3              & 16.3              \\
 CtrlGen \cite{hao2018controllable}   & 21.8           & 67.8              & 17.9              \\
 DPG \cite{gao2019disentangling}      & \textit{22.3}  & \textit{69.6}     & \textit{11.4}     \\
 \midrule
 SADM                         & \textbf{24.47} & \textbf{71.1}     & \textbf{10.9}     \\
 \bottomrule
\end{tabular}
\vspace{0.1cm}
\caption{Quantitative comparison in next-frame prediction on the KITTI Flow dataset.}
\vspace{-0.2cm}
\label{table:kitti_flow}
\end{table}

\begin{table}[h]
\setlength\tabcolsep{14.0pt}
\centering
\small
\begin{tabular}{cccc}
\toprule
\textbf{Method}      & t+1           & t+5           & t+9\\
\midrule
Repeat      & 67.1          & 52.1          & 38.3\\
S2S-dil \cite{luc2017predicting}     & -             & 59.4          & 47.8\\
PSPNet \cite{shahabeddin2018future}     & 71.3          & 60            & -   \\
Jin \cite{jin2017predicting}        & 66.1          & -             & -   \\
Terwilliger \cite{terwilliger2019recurrent} & 73.2          & 67.1          & 51.5\\
Bayes-WD-SL \cite{bhattacharyya2018bayesian} & \textbf{74.1}          & 65.1          & 51.2\\
F2MF-DN121 \cite{saric2020warp_s} & - & 69.6 & 57.9\\
\midrule
SADM & 73.8 & \textbf{70.3} & \textbf{60.1}\\
\bottomrule
\end{tabular}
\vspace{0.1cm}
\caption{Quantitative results of semantic map prediction on the Cityscapes dataset measured by the mIoU score.}
\vspace{-0.3cm}
\label{table:semantic-cityscape}
\end{table}

Following DPG \cite{gao2019disentangling}, we report the next-frame prediction results on the KITTI Flow dataset in Table \ref{table:kitti_flow}. 
For a fair comparison, we also include the commonly used Peak Signal-to-Noise Ratio (PSNR) and Structural Similarity Index Measure (SSIM) \cite{wang2004image} as the evaluation metrics, in addition to Learned Perceptual Image Patch Similarity (LPIPS) \cite{zhang2018unreasonable}. Our model improves DPG \cite{gao2019disentangling}, the previous state-of-the-art method evaluated on this dataset, by 9.7\% in terms of PSNR, 2.2\% in terms of SSIM, and 4.4\% in terms of LPIPS.

\textbf{Semantic segmentation mask prediction.}
We evaluate our model's performance on semantic mask prediction using the Cityscapes dataset, with the standard mean Intersection-over-Union score (mIoU) as the evaluation metric. Following \cite{luc2017predicting}, the scores are computed with respect to the ground-truth segmentation of the $20^{\text{th}}$ frame in each sequence. Table \ref{table:semantic-cityscape} shows the semantic mask prediction performance on multiple prediction lengths. Our method performs the best among the other methods, especially when the prediction horizon gets longer.

\subsection{Qualitative Results}

In Fig. \ref{fig:cityscape}, we compare to FVS \cite{wu2020future} and Seg2vid \cite{pan2019video}, two most recent methods that employ semantic segmentation or moving object segmentation to facilitate video prediction. 
The motivation of Seg2vid \cite{pan2019video} is that the high-level semantics of the scene will result in more accurate predictions.
However, without an explicit modeling, such as SADM, predicted flow fields from \cite{pan2019video} still suffer from over-smoothing. 
Moreover, given that most of the videos in Cityscapes are captured by a camera moving forward with a car, there is a strong tendency in the model from Seg2vid to produce flow fields showing zooming-in motion.
On the other hand, FVS \cite{wu2020future} separates the whole scene into moving and non-moving segments by moving object detection.
Although 2D affine transformations are predicted per frame to approximate the object motion, complex motion, including deformation and 3D rotation, may not be captured by a single 2D affine transformation. 
Even for non-moving rigid objects whose motion is induced by the camera motion, e.g., parked cars and buildings, their projected 2D flow fields still depend on the 3D geometries and thus are not 2D affine. 
With semantic-aware dynamics and inpainting, our model can generate high-quality video frames with more accurate motions.

\subsection{Ablation Study}
\label{section:ablation}

\begin{figure}[t]
    \centering
    \includegraphics[width=0.37\textwidth]{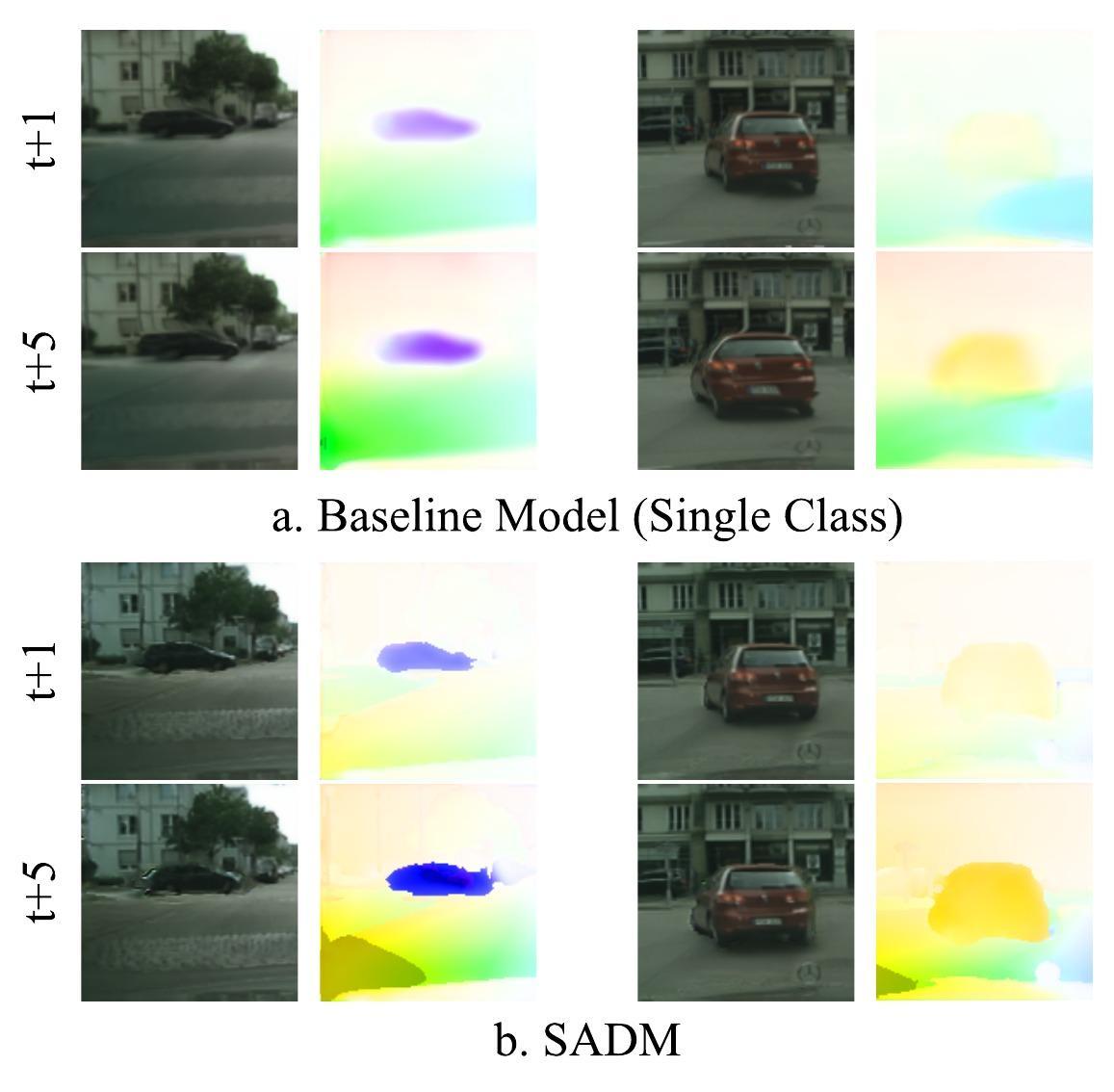}
    \caption{A baseline model without explicit modeling of the semantic-aware dynamics (single class) shows less accurate motion prediction than SADM and has more artifacts in the predicted video frames.}
    \vspace{-0.4cm}
    \label{ablation}
\end{figure}

To demonstrate the effectiveness of decomposing the video into semantically consistent regions for video prediction, we train a baseline model (``single class'' in Fig.~\ref{ablation}) with the same network architecture as SADM, and with a naive concatenation of semantic masks and flow fields as the input to the baseline model. 
The encoder and decoder of this baseline model share similar structures as those in SADM, besides that ordinary convolutions are used.
Note that an ordinary convolution layer has more parameters than a grouped convolutional layer ($O(K^2)$ v.s. $O(K)$). 
As shown in Fig. \ref{ablation}, without explicitly modeling the class-wise dynamics, the baseline model trained to predict flow fields or video frames has difficulties estimating the motion near the boundaries between different semantic regions (the black car in the top left). 
There is a heavy over-smoothing near the boundary of the car, which is problematic for the consecutive warping procedure, since the flow there will warp pixels on the car to the background or vice versa, generating the ``ghost effect''.
With the proposed semantic aware dynamic model, the motion of either the car or the background can be accurately estimated since the influence on motion estimation from occlusions is automatically handled through the decomposition. 
Similarly, in the bottom left of Fig. \ref{ablation}, the warped image using the flow predicted by the baseline model shows far more artifacts on the red car.

To show that the learned dynamics from our model is indeed semantic-aware, we test the model with semantic labels intentionally swapped in the input. 
For example, we input the ``car'' segments to the semantic-aware encoder that learns the dynamics of the ``road'' class, and vice versa.
As expected (Fig.~\ref{fig:permute}), the predicted motion of the ``car'' segments using the encoder for the ``road'' class (middle row) now looks like the one of the ``road'' (bottom row). 
Similarly, the predicted motion of the ``road'' segments using the encoder for the ``car'' class (middle row) now looks more like the one from the ``car'' (bottom row). 
This shows exactly that semantic-aware dynamics is captured by the proposed model.

\section{Discussion}

\begin{figure}[t]
    \centering
    \includegraphics[scale=0.55]{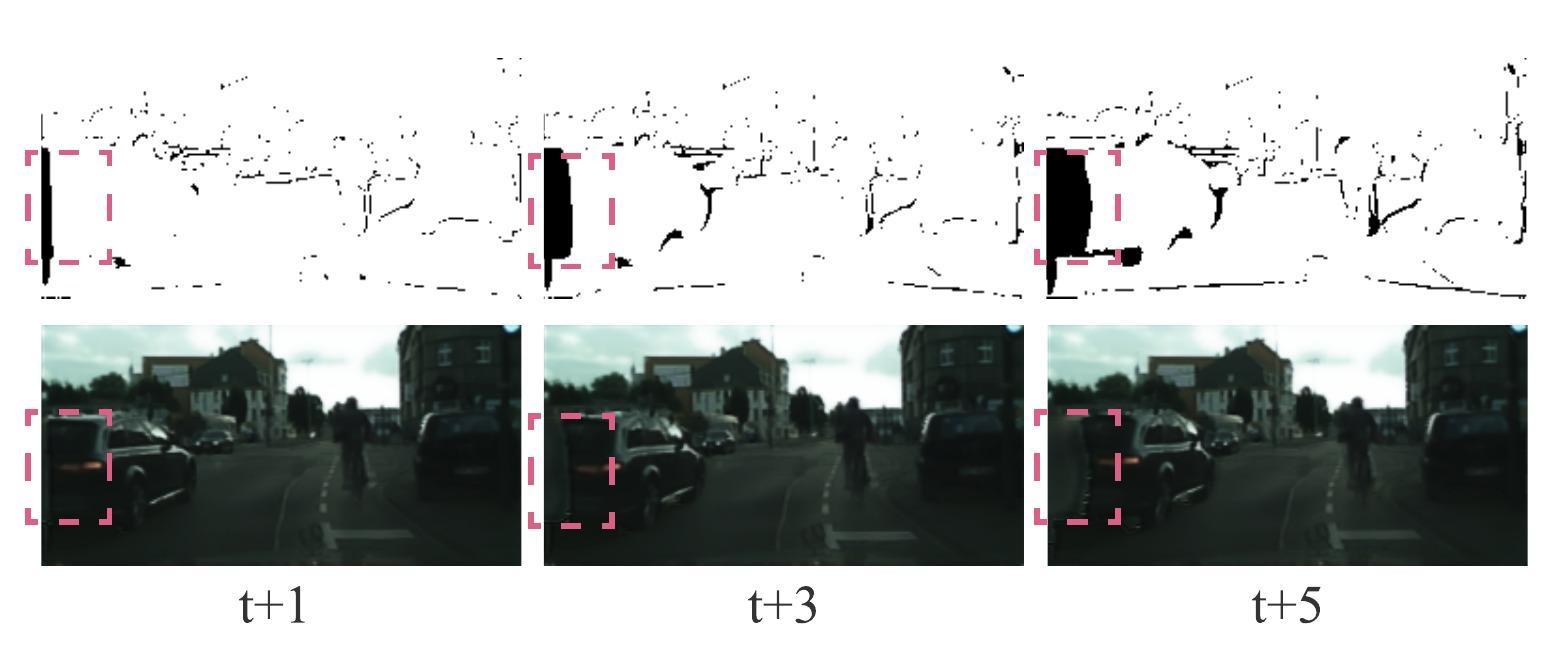}
    \caption{Complicated dis-occluded regions cause difficulties for the inpainting network, even if the estimated occlusions are accurate (top row).}
    \vspace{-0.3cm}
    \label{fig:failure}
\end{figure}

We have tested the hypothesis that representing object-level motion in a video can be beneficial for prediction. To that end, we have proposed a model that captures occlusions explicitly, and represents  class-specific motion. While such high-level modeling is beneficial to prediction, there are failure cases. Specifically, hallucinating 
the dis-occluded regions can lead to failure when the background is complex (Fig.  \ref{fig:failure}). 
As the time horizon grows, the prediction becomes 
increasingly unrealistic, as with other video prediction models, but the explicit modeling of objects and class-specific motion yields improvements over generic models. 
Also, we are constrained by classes for which we have training data, which limits  generalization. So, our work is only a first step to incorporate dynamical models that are informed by the semantics of objects in the scene, which we expect will ultimately facilitate intelligent interaction with physical scenes by autonomous agents.

\subsubsection*{Acknowledgment}

Research supported by ARL W911NF-20-1-0158 and ONR N00014-17-1-2072.

{\small
\bibliographystyle{ieee_fullname}
\bibliography{main}

\begin{thebibliography}{10}\itemsep=-1pt

\bibitem{babaeizadeh2017stochastic}
Mohammad Babaeizadeh, Chelsea Finn, Dumitru Erhan, Roy~H Campbell, and Sergey
  Levine.
\newblock Stochastic variational video prediction.
\newblock {\em arXiv preprint arXiv:1710.11252}, 2017.

\bibitem{bhattacharyya2018bayesian}
Apratim Bhattacharyya, Mario Fritz, and Bernt Schiele.
\newblock Bayesian prediction of future street scenes using synthetic
  likelihoods.
\newblock {\em arXiv preprint arXiv:1810.00746}, 2018.

\bibitem{brock2018large}
Andrew Brock, Jeff Donahue, and Karen Simonyan.
\newblock Large scale gan training for high fidelity natural image synthesis.
\newblock {\em arXiv preprint arXiv:1809.11096}, 2018.

\bibitem{byeon2018contextvp}
Wonmin Byeon, Qin Wang, Rupesh Kumar~Srivastava, and Petros Koumoutsakos.
\newblock Contextvp: Fully context-aware video prediction.
\newblock In {\em Proceedings of the European Conference on Computer Vision
  (ECCV)}, pages 753--769, 2018.

\bibitem{cai2018deep}
Haoye Cai, Chunyan Bai, Yu-Wing Tai, and Chi-Keung Tang.
\newblock Deep video generation, prediction and completion of human action
  sequences.
\newblock In {\em Proceedings of the European Conference on Computer Vision
  (ECCV)}, pages 366--382, 2018.

\bibitem{chen2018encoder}
Liang-Chieh Chen, Yukun Zhu, George Papandreou, Florian Schroff, and Hartwig
  Adam.
\newblock Encoder-decoder with atrous separable convolution for semantic image
  segmentation.
\newblock In {\em Proceedings of the European conference on computer vision
  (ECCV)}, pages 801--818, 2018.

\bibitem{Cordts2016Cityscapes}
Marius Cordts, Mohamed Omran, Sebastian Ramos, Timo Rehfeld, Markus Enzweiler,
  Rodrigo Benenson, Uwe Franke, Stefan Roth, and Bernt Schiele.
\newblock The cityscapes dataset for semantic urban scene understanding.
\newblock In {\em Proc. of the IEEE Conference on Computer Vision and Pattern
  Recognition (CVPR)}, 2016.

\bibitem{denton2018stochastic}
Emily Denton and Rob Fergus.
\newblock Stochastic video generation with a learned prior.
\newblock {\em arXiv preprint arXiv:1802.07687}, 2018.

\bibitem{gao2019disentangling}
Hang Gao, Huazhe Xu, Qi-Zhi Cai, Ruth Wang, Fisher Yu, and Trevor Darrell.
\newblock Disentangling propagation and generation for video prediction.
\newblock In {\em Proceedings of the IEEE International Conference on Computer
  Vision}, pages 9006--9015, 2019.

\bibitem{Geiger2013IJRR}
Andreas Geiger, Philip Lenz, Christoph Stiller, and Raquel Urtasun.
\newblock Vision meets robotics: The kitti dataset.
\newblock {\em International Journal of Robotics Research (IJRR)}, 2013.

\bibitem{Geiger2012CVPR}
Andreas Geiger, Philip Lenz, and Raquel Urtasun.
\newblock Are we ready for autonomous driving? the kitti vision benchmark
  suite.
\newblock In {\em Conference on Computer Vision and Pattern Recognition
  (CVPR)}, 2012.

\bibitem{hao2018controllable}
Zekun Hao, Xun Huang, and Serge Belongie.
\newblock Controllable video generation with sparse trajectories.
\newblock In {\em Proceedings of the IEEE Conference on Computer Vision and
  Pattern Recognition}, pages 7854--7863, 2018.

\bibitem{he2018probabilistic}
Jiawei He, Andreas Lehrmann, Joseph Marino, Greg Mori, and Leonid Sigal.
\newblock Probabilistic video generation using holistic attribute control.
\newblock In {\em Proceedings of the European Conference on Computer Vision
  (ECCV)}, pages 452--467, 2018.

\bibitem{hsieh2018learning}
Jun-Ting Hsieh, Bingbin Liu, De-An Huang, Li~F Fei-Fei, and Juan~Carlos
  Niebles.
\newblock Learning to decompose and disentangle representations for video
  prediction.
\newblock In {\em Advances in Neural Information Processing Systems}, pages
  517--526, 2018.

\bibitem{jia2016dynamic}
Xu Jia, Bert De~Brabandere, Tinne Tuytelaars, and Luc~V Gool.
\newblock Dynamic filter networks.
\newblock In {\em Advances in Neural Information Processing Systems}, pages
  667--675, 2016.

\bibitem{jin2020exploring}
Beibei Jin, Yu Hu, Qiankun Tang, Jingyu Niu, Zhiping Shi, Yinhe Han, and
  Xiaowei Li.
\newblock Exploring spatial-temporal multi-frequency analysis for high-fidelity
  and temporal-consistency video prediction.
\newblock In {\em Proceedings of the IEEE/CVF Conference on Computer Vision and
  Pattern Recognition}, pages 4554--4563, 2020.

\bibitem{jin2017predicting}
Xiaojie Jin, Huaxin Xiao, Xiaohui Shen, Jimei Yang, Zhe Lin, Yunpeng Chen,
  Zequn Jie, Jiashi Feng, and Shuicheng Yan.
\newblock Predicting scene parsing and motion dynamics in the future.
\newblock In {\em Advances in Neural Information Processing Systems}, pages
  6915--6924, 2017.

\bibitem{kim2019deep}
Dahun Kim, Sanghyun Woo, Joon-Young Lee, and In~So Kweon.
\newblock Deep video inpainting.
\newblock In {\em Proceedings of the IEEE Conference on Computer Vision and
  Pattern Recognition}, pages 5792--5801, 2019.

\bibitem{krizhevsky2017imagenet}
Alex Krizhevsky, Ilya Sutskever, and Geoffrey~E Hinton.
\newblock Imagenet classification with deep convolutional neural networks.
\newblock {\em Communications of the ACM}, 60(6):84--90, 2017.

\bibitem{li2018flow}
Yijun Li, Chen Fang, Jimei Yang, Zhaowen Wang, Xin Lu, and Ming-Hsuan Yang.
\newblock Flow-grounded spatial-temporal video prediction from still images.
\newblock In {\em Proceedings of the European Conference on Computer Vision
  (ECCV)}, pages 600--615, 2018.

\bibitem{liang2017dual}
Xiaodan Liang, Lisa Lee, Wei Dai, and Eric~P Xing.
\newblock Dual motion gan for future-flow embedded video prediction.
\newblock In {\em Proceedings of the IEEE International Conference on Computer
  Vision}, pages 1744--1752, 2017.

\bibitem{liu2018image}
Guilin Liu, Fitsum~A Reda, Kevin~J Shih, Ting-Chun Wang, Andrew Tao, and Bryan
  Catanzaro.
\newblock Image inpainting for irregular holes using partial convolutions.
\newblock In {\em Proceedings of the European Conference on Computer Vision
  (ECCV)}, pages 85--100, 2018.

\bibitem{liu2017video}
Ziwei Liu, Raymond~A Yeh, Xiaoou Tang, Yiming Liu, and Aseem Agarwala.
\newblock Video frame synthesis using deep voxel flow.
\newblock In {\em Proceedings of the IEEE International Conference on Computer
  Vision}, pages 4463--4471, 2017.

\bibitem{lotter2016deep}
William Lotter, Gabriel Kreiman, and David Cox.
\newblock Deep predictive coding networks for video prediction and unsupervised
  learning.
\newblock {\em arXiv preprint arXiv:1605.08104}, 2016.

\bibitem{luc2017predicting}
Pauline Luc, Natalia Neverova, Camille Couprie, Jakob Verbeek, and Yann LeCun.
\newblock Predicting deeper into the future of semantic segmentation.
\newblock In {\em Proceedings of the IEEE International Conference on Computer
  Vision}, pages 648--657, 2017.

\bibitem{luo2017unsupervised}
Zelun Luo, Boya Peng, De-An Huang, Alexandre Alahi, and Li Fei-Fei.
\newblock Unsupervised learning of long-term motion dynamics for videos.
\newblock In {\em Proceedings of the IEEE Conference on Computer Vision and
  Pattern Recognition}, pages 2203--2212, 2017.

\bibitem{marwah2017attentive}
Tanya Marwah, Gaurav Mittal, and Vineeth~N Balasubramanian.
\newblock Attentive semantic video generation using captions.
\newblock In {\em Proceedings of the IEEE International Conference on Computer
  Vision}, pages 1426--1434, 2017.

\bibitem{pan2019video}
Junting Pan, Chengyu Wang, Xu Jia, Jing Shao, Lu Sheng, Junjie Yan, and
  Xiaogang Wang.
\newblock Video generation from single semantic label map.
\newblock In {\em Proceedings of the IEEE Conference on Computer Vision and
  Pattern Recognition}, pages 3733--3742, 2019.

\bibitem{patwardhan2007video}
Kedar~A Patwardhan, Guillermo Sapiro, and Marcelo Bertalm{\'\i}o.
\newblock Video inpainting under constrained camera motion.
\newblock {\em IEEE Transactions on Image Processing}, 16(2):545--553, 2007.

\bibitem{pintea2014deja}
Silvia~L Pintea, Jan~C van Gemert, and Arnold~WM Smeulders.
\newblock D{\'e}ja vu.
\newblock In {\em European Conference on Computer Vision}, pages 172--187.
  Springer, 2014.

\bibitem{reda2018sdc}
Fitsum~A Reda, Guilin Liu, Kevin~J Shih, Robert Kirby, Jon Barker, David
  Tarjan, Andrew Tao, and Bryan Catanzaro.
\newblock Sdc-net: Video prediction using spatially-displaced convolution.
\newblock In {\em Proceedings of the European Conference on Computer Vision
  (ECCV)}, pages 718--733, 2018.

\bibitem{reed2016generative}
Scott Reed, Zeynep Akata, Xinchen Yan, Lajanugen Logeswaran, Bernt Schiele, and
  Honglak Lee.
\newblock Generative adversarial text to image synthesis.
\newblock {\em arXiv preprint arXiv:1605.05396}, 2016.

\bibitem{ronneberger2015u}
Olaf Ronneberger, Philipp Fischer, and Thomas Brox.
\newblock U-net: Convolutional networks for biomedical image segmentation.
\newblock In {\em International Conference on Medical image computing and
  computer-assisted intervention}, pages 234--241. Springer, 2015.

\bibitem{saric2020warp_s}
Josip S.~et al.
\newblock Warp to the future.
\newblock In {\em CVPR}, 2020.

\bibitem{saito2017temporal}
Masaki Saito, Eiichi Matsumoto, and Shunta Saito.
\newblock Temporal generative adversarial nets with singular value clipping.
\newblock In {\em Proceedings of the IEEE International Conference on Computer
  Vision}, pages 2830--2839, 2017.

\bibitem{shahabeddin2018future}
Seyed shahabeddin Nabavi, Mrigank Rochan, and Yang Wang.
\newblock Future semantic segmentation with convolutional lstm.
\newblock In {\em BMVC}, page 137, 2018.

\bibitem{sun2018pwc}
Deqing Sun, Xiaodong Yang, Ming-Yu Liu, and Jan Kautz.
\newblock Pwc-net: Cnns for optical flow using pyramid, warping, and cost
  volume.
\newblock In {\em Proceedings of the IEEE Conference on Computer Vision and
  Pattern Recognition}, pages 8934--8943, 2018.

\bibitem{terwilliger2019recurrent}
Adam Terwilliger, Garrick Brazil, and Xiaoming Liu.
\newblock Recurrent flow-guided semantic forecasting.
\newblock In {\em 2019 IEEE Winter Conference on Applications of Computer
  Vision (WACV)}, pages 1703--1712. IEEE, 2019.

\bibitem{tulyakov2018mocogan}
Sergey Tulyakov, Ming-Yu Liu, Xiaodong Yang, and Jan Kautz.
\newblock Mocogan: Decomposing motion and content for video generation.
\newblock In {\em Proceedings of the IEEE conference on computer vision and
  pattern recognition}, pages 1526--1535, 2018.

\bibitem{villegas2017decomposing}
Ruben Villegas, Jimei Yang, Seunghoon Hong, Xunyu Lin, and Honglak Lee.
\newblock Decomposing motion and content for natural video sequence prediction.
\newblock {\em arXiv preprint arXiv:1706.08033}, 2017.

\bibitem{villegas2017learning}
Ruben Villegas, Jimei Yang, Yuliang Zou, Sungryull Sohn, Xunyu Lin, and Honglak
  Lee.
\newblock Learning to generate long-term future via hierarchical prediction.
\newblock In {\em Proceedings of the 34th International Conference on Machine
  Learning-Volume 70}, pages 3560--3569. JMLR. org, 2017.

\bibitem{vondrick2016generating}
Carl Vondrick, Hamed Pirsiavash, and Antonio Torralba.
\newblock Generating videos with scene dynamics.
\newblock In {\em Advances In Neural Information Processing Systems}, pages
  613--621, 2016.

\bibitem{walker2017pose}
Jacob Walker, Kenneth Marino, Abhinav Gupta, and Martial Hebert.
\newblock The pose knows: Video forecasting by generating pose futures.
\newblock In {\em Proceedings of the IEEE International Conference on Computer
  Vision}, pages 3332--3341, 2017.

\bibitem{wang2018video}
Ting-Chun Wang, Ming-Yu Liu, Jun-Yan Zhu, Guilin Liu, Andrew Tao, Jan Kautz,
  and Bryan Catanzaro.
\newblock Video-to-video synthesis.
\newblock {\em arXiv preprint arXiv:1808.06601}, 2018.

\bibitem{wang2018high}
Ting-Chun Wang, Ming-Yu Liu, Jun-Yan Zhu, Andrew Tao, Jan Kautz, and Bryan
  Catanzaro.
\newblock High-resolution image synthesis and semantic manipulation with
  conditional gans.
\newblock In {\em Proceedings of the IEEE conference on computer vision and
  pattern recognition}, pages 8798--8807, 2018.

\bibitem{wang2004image}
Zhou Wang, Alan~C Bovik, Hamid~R Sheikh, and Eero~P Simoncelli.
\newblock Image quality assessment: from error visibility to structural
  similarity.
\newblock {\em IEEE transactions on image processing}, 13(4):600--612, 2004.

\bibitem{wang2003multiscale}
Zhou Wang, Eero~P Simoncelli, and Alan~C Bovik.
\newblock Multiscale structural similarity for image quality assessment.
\newblock In {\em The Thrity-Seventh Asilomar Conference on Signals, Systems \&
  Computers, 2003}, volume~2, pages 1398--1402. Ieee, 2003.

\bibitem{wichers2018hierarchical}
Nevan Wichers, Ruben Villegas, Dumitru Erhan, and Honglak Lee.
\newblock Hierarchical long-term video prediction without supervision.
\newblock {\em arXiv preprint arXiv:1806.04768}, 2018.

\bibitem{wu2020future}
Yue Wu, Rongrong Gao, Jaesik Park, and Qifeng Chen.
\newblock Future video synthesis with object motion prediction.
\newblock In {\em Proceedings of the IEEE/CVF Conference on Computer Vision and
  Pattern Recognition}, pages 5539--5548, 2020.

\bibitem{xingjian2015convolutional}
SHI Xingjian, Zhourong Chen, Hao Wang, Dit-Yan Yeung, Wai-Kin Wong, and
  Wang-chun Woo.
\newblock Convolutional lstm network: A machine learning approach for
  precipitation nowcasting.
\newblock In {\em Advances in neural information processing systems}, pages
  802--810, 2015.

\bibitem{xu2018structure}
Jingwei Xu, Bingbing Ni, Zefan Li, Shuo Cheng, and Xiaokang Yang.
\newblock Structure preserving video prediction.
\newblock In {\em Proceedings of the IEEE Conference on Computer Vision and
  Pattern Recognition}, pages 1460--1469, 2018.

\bibitem{xu2019deep}
Rui Xu, Xiaoxiao Li, Bolei Zhou, and Chen~Change Loy.
\newblock Deep flow-guided video inpainting.
\newblock {\em arXiv preprint arXiv:1905.02884}, 2019.

\bibitem{xue2016visual}
Tianfan Xue, Jiajun Wu, Katherine Bouman, and Bill Freeman.
\newblock Visual dynamics: Probabilistic future frame synthesis via cross
  convolutional networks.
\newblock In {\em Advances in neural information processing systems}, pages
  91--99, 2016.

\bibitem{yang2018pose}
Ceyuan Yang, Zhe Wang, Xinge Zhu, Chen Huang, Jianping Shi, and Dahua Lin.
\newblock Pose guided human video generation.
\newblock In {\em Proceedings of the European Conference on Computer Vision
  (ECCV)}, pages 201--216, 2018.

\bibitem{yang2020learning}
Yanchao Yang, Yutong Chen, and Stefano Soatto.
\newblock Learning to manipulate individual objects in an image.
\newblock In {\em Proceedings of the IEEE/CVF Conference on Computer Vision and
  Pattern Recognition}, pages 6558--6567, 2020.

\bibitem{yang2019unsupervised}
Yanchao Yang, Antonio Loquercio, Davide Scaramuzza, and Stefano Soatto.
\newblock Unsupervised moving object detection via contextual information
  separation.
\newblock In {\em Proceedings of the IEEE/CVF Conference on Computer Vision and
  Pattern Recognition}, pages 879--888, 2019.

\bibitem{zhang2018unreasonable}
Richard Zhang, Phillip Isola, Alexei~A Efros, Eli Shechtman, and Oliver Wang.
\newblock The unreasonable effectiveness of deep features as a perceptual
  metric.
\newblock In {\em Proceedings of the IEEE Conference on Computer Vision and
  Pattern Recognition}, pages 586--595, 2018.

\bibitem{zhao2018learning}
Long Zhao, Xi Peng, Yu Tian, Mubbasir Kapadia, and Dimitris Metaxas.
\newblock Learning to forecast and refine residual motion for image-to-video
  generation.
\newblock In {\em Proceedings of the European Conference on Computer Vision
  (ECCV)}, pages 387--403, 2018.

\bibitem{CycleGAN2017}
Jun-Yan Zhu, Taesung Park, Phillip Isola, and Alexei~A Efros.
\newblock Unpaired image-to-image translation using cycle-consistent
  adversarial networks.
\newblock In {\em Computer Vision (ICCV), 2017 IEEE International Conference
  on}, 2017.

\end{thebibliography}
}

\newpage
\onecolumn


\section{Supplementary Material}

\subsection{Training Weights}
\label{sec:supp_weights}

For training the stochastic semantic aware dynamic model (Eq. \ref{eq:vae-loss}):
\begin{equation}
    \mathcal{L}_{\text {dynamic}} = \mathcal{L}_f + \mathcal{L}_m + \beta \mathcal{L}_{kl}
\end{equation}
we set $\beta = 0.1$.
For training the semantic aware inpainting network (dis-occlusion synthesis, Eq. \ref{eq:inpainting-loss}):
\begin{equation}
\begin{aligned}
    \mathcal{L}_{\varphi} = \sum_{t=T+1}^{T+K} (1-\Omega_d^t)\cdot\|\Tilde{x}^t - \hat{x}^t\|_{1} + 
    \lambda \mathcal{L}_{per}( \Tilde{x}^t, x^t)
    + \gamma \sum_{t=T+1}^{T+K}  D_x(\Tilde{x}^t,\Tilde{m}^t) + \eta D_v(\{ \Tilde{x}^t \}, \{\Tilde{m}^t\}) 
\end{aligned}
\end{equation}
we set $\lambda = 2.0$, $\gamma = 2.0$ and $\eta = 1.0$.

\subsection{Ablation Study with Different Number of Semantic Classes on the Cityscapes Dataset}
\label{sec:supp_ablation}

We have experimented with different numbers of classes ($C$) for training our semantic-aware dynamic model.
Our observations are: 1) the performance of our model, measured in terms of the video prediction quality, is robust with respect to the number of classes $C$; 2) the optimal performance may not be achieved by the model which has access to the full range of semantic classes;
The results we reported in the main paper are from our model trained with nine classes ($C=9$), i.e., the 20 classes within the Cityscapes dataset include 12 static categories and 8 moving categories, and we merge the 12 static categories into a single class, resulting in 9 classes.
In Table \ref{table:supp_ablation}, we show the quantitative results of SADM trained with $C=9$ and $C=20$ (no merging of any classes), together with comparisons to other state-of-the-art methods on the Cityscapes dataset.
One can observe that both of them achieve comparable or better performance than the other state of the arts. And SADM with $C=9$ performs slightly better than SADM with $C=20$.

\begin{table}[h]
\setlength{\tabcolsep}{4pt}
\centering
\begin{tabular}{ccccc}
\toprule
    & \multicolumn{2}{c}{\textbf{MS-SSIM} ($\times$1e-2) $\uparrow$} & \multicolumn{2}{c}{\textbf{LPIPS} ($\times$1e-2) $\downarrow$} \\
    \textbf{Method} & t+1 & t+5 & t+1 & t+5                \\
    \midrule
    PredNet \cite{lotter2016deep} & 84.03 & 75.21             & 25.99             & 36.03             \\
    MCNET \cite{villegas2017decomposing}      & 89.69             & 70.58             & 18.88             & 37.34             \\
    Voxel Flow \cite{liu2017video} & 83.85             & 71.11             & 17.37             & 28.79             \\
    Vid2vid \cite{wang2018video}    & 88.16             & 75.13             & 10.58             & 20.14             \\
    Seg2vid \cite{pan2019video}    & 88.32             & 61.63             & 9.69             & 25.99             \\
    FVS \cite{wu2020future}        & 89.10             & 75.68             & 8.50              & 16.50              \\
    \midrule
    SADM-20 & 93.25 & 79.14 & 10.06 & 17.55\\
    SADM-9 & \textbf{95.99}    & \textbf{83.51}    & \textbf{7.67}    & \textbf{14.93}    \\
    \bottomrule
\end{tabular}
\vspace{0.1cm}
\caption{Ablation study on the number of classes used for our semantic-aware dynamic model with comparisons to other state-of-the-art methods on the Cityscapes dataset.}
\label{table:supp_ablation}
\end{table}

Our conjecture is that the estimation error in the semantic maps generated by Deeplab or any off-the-shelf semantic segmentation networks have a larger impact on SADM with $C=20$. For example, the segmentation boundary may not be accurate between two static classes and the errors could propagate. However, when these two static classes are merged, the in-between erroneous boundaries just disappear, leaving no errors for propagation, which indeed can facilitate the training.



\subsection{Efficiency}

As mentioned in Sec. \ref{sec:sadm}, we use grouped convolution in the encoder/decoder and employ class-specific recurrent units. 
This breaks combinatorial complexity, can run in parallel and leaves the modeling of relations among classes to a separate fusion layer besides the context-aware MLP in the decoders. 
As discussed in Sec. \ref{section:ablation}, jointly modeling the dynamics of all classes would require much larger capacity and a much larger training set.

To provide a quantitative measurement of the complexity of our model, we compare the number of parameters w.r.t to several state-of-art models as in Tab. \ref{tab:nparam}, from which we can make the following observations:
1) even with more parameters, SADM with single class ($C=1$) still can not learn the dynamics well;
2) the number of parameters in SADMs with $C=9$ and $C=20$ are comparable to the other state-of-the-art methods, which demonstrates that the learning efficiency of SADM does not come from any increase in the network capacity but purely from the explicit modeling of semantic-aware dynamics.

\begin{table}[h]
\small
\begin{tabular}{cccccc|ccc}
\midrule
\textbf{Method}                       & PredNet \cite{lotter2016deep} & ContextVP \cite{byeon2018contextvp} & DVF \cite{liu2017video}  & STMFA \cite{jin2020exploring} & Seg2vid \cite{pan2019video} & \multicolumn{1}{l}{SADM-9} & SADM-20 & SADM-ablation \\
\multicolumn{1}{l}{\textbf{\#param}} & 6.9M    & 8.6M       & 8.9M & 7.6M     & 202M    & 8.6M                        & 23.2M   & 38.1M            \\ \midrule
\end{tabular}
\caption{Number of parameters in comparison with state-of-the-art models.}
\label{tab:nparam}
\end{table}

\begin{figure}[h]
    \centering
    \includegraphics[scale=0.8]{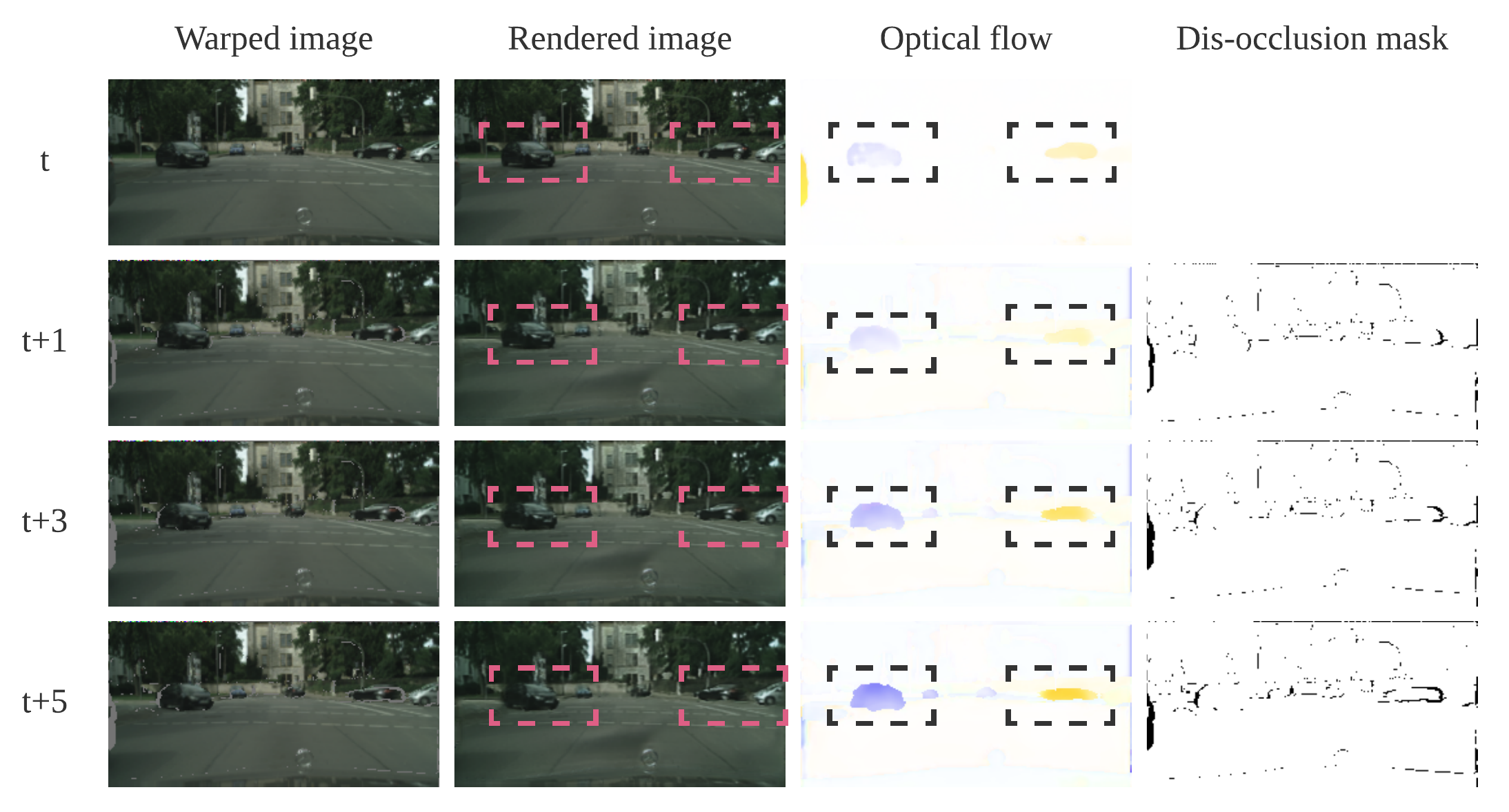}
    \caption{Visualization of two ``cars'' moving in different directions}
    \label{fig:two_car}
\end{figure}

\subsection{Why Not Instance Segmentation?}
 
First, instance-level annotation is prohibitively expensive.
While instance-level modeling is in principle more accurate, instance resolution engenders errors that can cause the quality of the prediction to degrade beyond the coarser granularity provided by semantic layers. Tab. \ref{table:supp_ablation} show ablation studies, whereby modeling more classes does not necessarily result in improvement due to estimation/segmentation errors (one can consider the extreme case of our model where each instance is a class). 

Moreover, even though our model is not instance-specific, it can actually capture the dynamics of instances within a class (manifested in the flow field), due to our explicit modeling of class-specific dynamics (in Fig. \ref{fig:two_car}).

\end{document}